# Implementation Of Back-Propagation Neural Network For Isolated Bangla Speech Recognition

Md. Ali Hossain[1], Md. Mijanur Rahman[2], Uzzal Kumar Prodhan[3], Md. Farukuzzaman Khan[4]

[1]Dept. of Computer Science & Engineering, Bangladesh University, Bangladesh.
`ali.cse.bd@gmail.com`

[2,3]Dept. of Computer Science & Engineering, Jatiya Kabi Kazi Nazrul Islam University, Bangladesh.
`mijan_cse@yahoo.com`[2] and `uzzal_bagerhat@yahoo.com`[3]

[4]Dept. of Computer Science & Engineering, Islamic University, Bangladesh.
`mfkhanbd2@gmail.com`

## ABSTRACT

*This paper is concerned with the development of Back-propagation Neural Network for Bangla Speech Recognition. In this paper, ten bangla digits were recorded from ten speakers and have been recognized. The features of these speech digits were extracted by the method of Mel Frequency Cepstral Coefficient (MFCC) analysis. The mfcc features of five speakers were used to train the network with Back propagation algorithm. The mfcc features of ten bangla digit speeches, from 0 to 9, of another five speakers were used to test the system. All the methods and algorithms used in this research were implemented using the features of Turbo C and C++ languages. From our investigation it is seen that the developed system can successfully encode and analyze the mfcc features of the speech signal to recognition. The developed system achieved recognition rate about 96.332% for known speakers (i.e., speaker dependent) and 92% for unknown speakers (i.e., speaker independent).*

## 1. Introduction

To communicate with each other, Speech is probably the most efficient way. It is possible to use speech as a useful interface to interact with machines[1]. Speech recognition research work has been started since 1930. However Bangla speech recognition research work has been started since around 2000 [2]. Besides English language there is a lot of research experiment and achieved result in various languages throughout the world. But in Bangla language, early researchers in this field have limited success in phonemes [3,4], letters [5], words [6] or small vocabulary continuous speech[7] for single speaker. In our system, we have captured speech from ten different speakers, which may an early attempt for developing speaker independent isolated Bangla digit speech recognition system in Bangla language. With a rich heritage, Bangla is an important language . It is spoken by approximately 8% of the world population [8]. But, for the computerization of this language, a systematic and scientific effort has not started yet. To support rapidly developing computerization of Bangla Language, one of the most important issues is Bangla speech recognition. Accordingly, we have developed a Bangla speech recognition system. A Neural Network is an information Processing Paradigm and it is stimulated by the way biological nervous systems, like the brain process Information[9]. Simple computational elements operating in parallel are included in Neural networks [1]. The network function is determined largely by the connections between elements. A neural network can be trained so that a particular input guides to a specific target output [1]. Neural network can be used in different sector. There are many applications of NNs limited only by our thoughts. Innovation is a solution to success, so dude use NNs to generate something which will modernize the world! For the sake of writing, a few Neural Networks applications are in Speech Recognition, Optical Character Recognition (OCR), Modelling human behaviour, Classification of patterns, Loan risk analysis, music generation, Image analysis, Creating new art forms, Stock market prediction, etc. In our research work, Multilayer Feed-forward Network with Back-propagation algorithm is used to recognize isolated Bangla speech digits from 0 to 9. All the methods and algorithms discussed in this paper were implemented using the features of C and C++ languages.

## 2. Feed forward Networks

The simplest type of feed-forward network that use supervised learning is Perceptron. Binary threshold units arranged into layers made a perceptron shown in Figure-1[10]. It is trained by the Delta Rule or variations thereof. The Delta Rule can be applied directly for the case of a single layer perceptron, as shown in Figure-1(a). For being perceptron's activations are binary, reduces this general learning rule to the Perceptron Learning Rule. This rules says, if an input is $y_i = 1$ (Active) and the output $y_j$ is incorrect , then depending on the desired output , the weight $w_{ji}$ need to be either decreased(if desired output is 0) or increased(if desired output is 1) by a small amount ε [10]. To find a set of weights to accurately organize the patterns in any training set, This procedure is insured, if the patterns are linearly distinguishable ( i.e. by a straight line, they can be split into two classes). However, most training sets are not linearly distinguishable (For instance, the simple XOR function); for these cases we require multiple layers.

Figure-1(b) shows *Multi-layer perceptrons* (MLPs). It can theoretically learn any function, but they are very complex to train. It is not possible to apply The Delta Rule directly to MLPs because in the hidden layer, there are no targets. If an MLP don't use discrete activation functions (i.e., threshold functions) that is it uses continuous activation functions (i.e., sigmoids functions), then it turn out to be possible to use partial derivatives and the chain rule .These rules is used to derive the influence of any weight on any output activation, that indicates, to reduce the network's error, how to modify that weight [10]. This generalized Delta Rule is known as back-propagation.

Any number of hidden layers can have in MLPs, although for many applications, a single hidden layer is sufficient, and more hidden layers tend to make training slower, for this reason the terrain in weight space becomes more complicated. There are many ways that MLPs can also be architecturally constrained, such as by limiting the weights values, or by limiting their geometrically local areas connectivity, or tying different weights together.

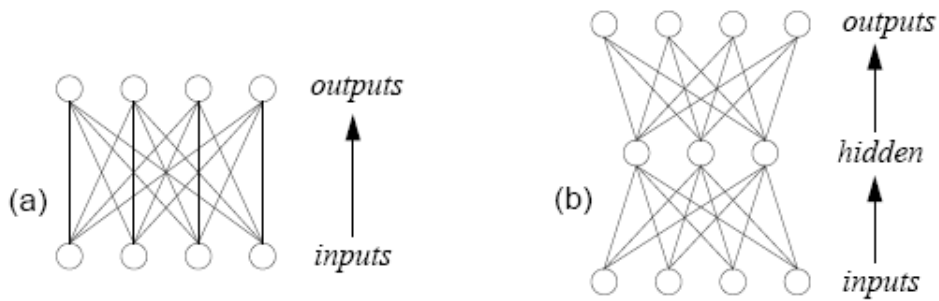

Figure 1. Perceptrons. (a) Single layer perceptron; (b) multi-layer perceptron.

## 2.1 Generalized Delta Rule (Back-Propagation)

The generalized delta rule is a supervised learning algorithm [11]. It is used to train a multilayer neural network that maps the relation between the target output and actual output. During the training period, the input pattern is passed through the network with network connection weights and biases of the activation or transfer functions. Initially, the values and biases are assigned by a small random numbers. In this rule, repeatedly presenting input-output pairs and then modifying weights; the modification of weight reduces the network error. The training of the network proceeds as follows:

- First, the input pattern is assigned to input neurons; and send to hidden neurons with weights and computed its activation; and then send this to the output neurons with weights and computed its activation, which represent the network's output response of the input pattern.
- Second, the output responses are compared to the desired output and an error term is computed.
- Third, the error information is used to update the network weights and biases. The modification of weight is computed by the four different parameters: a learning rate, the derivative of activation function, the error term, and the current activity at the input layer.
- Fourth, each hidden unit computes its error. This is done with an output unit's error, which is sending it backwards as a signal to a hidden unit.
- Fifth, after computing the hidden unit error, the weights of the input-to-hidden units are updated using the same equation that was used in the output layer.

If there is more than one layer of hidden units, then this procedure can be repeated iteratively. That is, each hidden unit error in one layer, as an error signal, can be propagated backwards to an adjacent layer once the hidden unit weights have been modified. Then the next training pattern can be presented to the input units, and the learning process occurs again [12].

## 2.2 Working with back-propagation Algorithm

The application of the generalized delta rule involves two phases. At the first phase, to compute the output values $y_p$ o for each output unit , the input x is presented and propagated forward through the network. Desired output value do is compared with this output, for each output unit resulting in an error signal $\delta_p$ o. To calculate the appropriate weight changes, a backward pass is involved by the second phase through the network during which the error signal is passed to each unit in the network [13]. Figure-2 shows the flowchart of the training of the neural network with Back-propagation algorithm.

***Weight adjustments with sigmoid activation function:***

The weight of a connection is adjusted by an amount proportional to the product of an error signal δ, on the unit k receiving the input and the output of the unit j sending this signal along the

connection:
$$\Delta_p w_{jk} = \gamma\, \delta^P_h\, y^P_h$$

- The error signal for an output unit is given by: $\delta^Q_o = (d^Q_o - y^Q_o)\, \mathcal{F'}(s^Q_o)$

- If the activation function F, the 'sigmoid' function defined as:

$$y^p = \mathcal{F}(S^P) = \frac{1}{1+e^{-S^P}}$$

- In this case the derivative is equal to

$$\mathcal{F}'(S^P) = \frac{\partial}{\partial S^P}\frac{1}{1+e^{-S^P}}$$
$$= \frac{1}{(1+e^{-S^P})2}(-e^{-S^P})$$
$$= \frac{1}{(1+e^{-S^P})2}\frac{e^{-S^P}}{(1+e^{-S^P})}$$
$$= y^P(1- y^P).$$

- Such that the output unit error signal can be Written as:

$$\delta^P_o = (d^P_o - y^P_o)\, y^P_o(1- y^P_o).$$

- The error signal for a hidden unit is determined recursively in terms of error signals of the units to which it directly connects and the weights of those connections. For the sigmoid activation function:

$$\delta^P_h = \mathcal{F}'(S^P_h)\sum_{o=1}^{No}\delta^P_o\, w_{ho} = y^P_h(1- y^P_h)\sum_{o=1}^{No}\delta^P_o\, w_{ho}$$

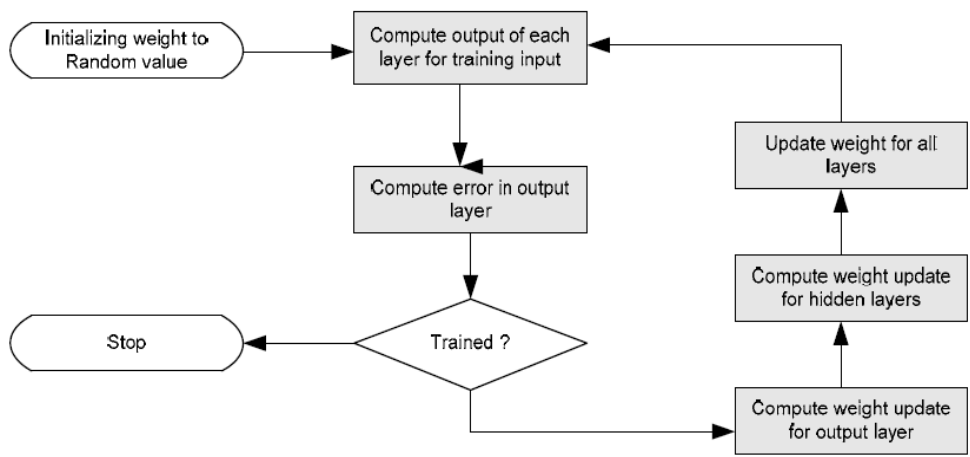

Figure 2. Back-propagation training algorithm

# 3. IMPLEMENTATION

## 3.1 Speech Data Acquisition

In a sound proof laboratory environment, Bangla speech words recording was completed with the help of close-talking microphone, sound recorder software and high quality sound card. To represent a signal, Wave form is the most general way [14-16]. To make a sample database, The 10 Bangla digits originated from ten speakers were recorded as *wav* file. At a sampling rate of 8.00 KHz and coded in 8 bits PCM [17], The utterances were recorded. At first, we discard 58 bytes (file header) from the beginning of the wave file for extracting wave data. After then wave data read as character. Required voiced data is extracted from the input speech ( may containing unvoice, voice and silence signal) by the data extraction process. By using the start and end-point detection technique [8,18], the voiced data is extracted from the speech file and stored in a text file as integer data.

## 3.2 Feature Extraction

The most important part of all recognition systems is the feature extraction that translates the speech signal to some digital form having meaningful features. Obviously, for any recognition system, a good feature may produce a good result. Frame blocking, Preemphasis, Windowing and the computation of Mel Frequency Cepstrum Coefficient (MFCC) are the some signal processing steps included in Feature extraction process, as shown in Figure-3 [8,19]. Initially, each speech word was segmented into frame (a set of samples). A frame represent typically 16 to 32 ms of speech. Then apply the further computation on these speech frames. In our research work, Hamming window function is used for feature extraction, which used in speech recognition as one of the most popular windows. To extract a set of features that represents Mel Frequency Cepstrum Coefficients (MFCC) of the signal, The pre-processed and windowed speech signal is then passed through some computational steps. The Discrete Fourier Transform (DFT), computation of first two formant frequencies, Mel frequency warping, Discrete Cosine Transform (DCT) and finally the computation of Mel Frequency Cepstrum Coefficient (MFCC) are included in the computation steps of MFCC, as shown in Figure-3 [8, 19-21].

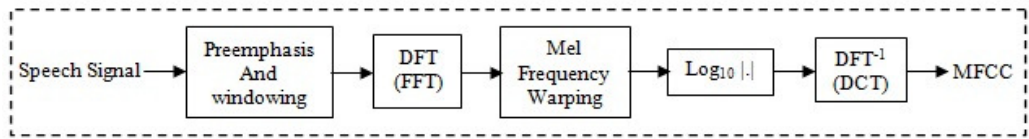

Figure 3. Computation of MFCC features.

## 3.3 Training and Testing the Network

In the past, Neural networks were used by many authors for speech recognition [22-25]. For our implementation, Multilayer Feed forward Network (three layer neural network) has been created using C & C++. For training the network, Back Propagation algorithm was used. The network consists of an input layer of 250 neurons, one hidden layer of 16 neurons and an output layer contains 10 neurons used to recognize 10 speech words. We used a set of 250 mfcc (Mel Frequency Cepstrum coefficients) feature values as input pattern for the neural network. The input values are in a range of -7 up to 1.5. In our design we put all these input values in a 'Input Layer' variable matrix. Linear activation function elements are included in the output layer. As 10 output neurons, output matrix contains 10x10 unit matrix. Non-linear sigmoidal activation function is included in the hidden layer. For training the network, at first Randomize weights and biases are set using random function ranging values from -1 to 1. Last layer does not require weights and first layer does not need biases. For each training pattern, network layers weights and biases are updated using back-propagation algorithms to reach the target, with the following equations.

**For output layer:**

Delta = (TargetO - ActualO) * ActualO * (1 - ActualO)
Weight = Weight + LearningRate * Delta * Input

**For hidden layers:**

Delta = ActualO * (1-ActualO) * Summation(Weight_from_current_to_next AND Delta_of_next)
Weight = Weight + LearningRate * Delta * Input

Where, TargetO= Target Output (This outputs are set to recognize ten difference words )
ActualO= Actual Output (Calculated output for training phase)
Delta= Bias of Neuron
LearningRate = Learning Rate to Update network weight.

A set of 15 mfcc features pattern of each speech word are used as training pattern to reach its target. The weights and biases of the network are updated until the network error reaches almost zero. At testing phase, the trained network was simulated with unknown speech pattern. It was observed that the trained network performs very well and more than ten words can be recognized by using the developed system.

## 4. Results and Discussion

In this experiment, 300 samples of ten Bangla digits, from 0 to 9, (i.e., 30 samples for each digit) were recorded from ten speakers and then extracted mfcc features from these Bangla speech words. The MFCC features were used with 8-mfcc/frame. 150 mfcc features of speech words were used for training the network and another 150 mfcc features were used for testing the network in the recognition phase. The developed system achieved 96.332% recognition accuracy for known speakers (i.e., speaker dependent) and 92% accuracy for unknown speakers (i.e., speaker independent). The detailed results are shown in Table-1.

In this paper, an attempt was made to develop Back-propagation Neural Network for Isolated Bangla Speech Recognition system. This is a single word recognition system, where ten Bangla digits are recognized at a time. All the test patterns were conducted with ten different Speakers from different age. Where fives are trained speakers and fives are test speakers. Instruments should have constant settings. The speaker was a 25-aged male person. Performance with some speakers from different age group was also tested. It was observed that speaking habit or style of speaker affects the performance of the system, so the speakers should be well trained. The sources of errors also include speeds of utterance and loudness variation. Also, the characteristics of microphone and other recording instruments and background noises affect the system performance.

| Bangla Digit | No. of samples for testing | Speaker Dependent (Known speaker's speech were used for testing) | | Speaker Independent (Unknown speaker's speech were used for testing) | |
|---|---|---|---|---|---|
| | | No. of Properly Recognized Digit | Recognition Rate (%) | No. of Properly Recognized Digit | Recognition Rate (%) |
| 0 | 15 | 14 | 93.33 | 14 | 93.33 |
| 1 | 15 | 15 | 100 | 14 | 93.33 |
| 2 | 15 | 14 | 93.33 | 13 | 86.67 |
| 3 | 15 | 15 | 100 | 14 | 93.33 |
| 4 | 15 | 15 | 100 | 14 | 93.33 |
| 5 | 15 | 14 | 93.33 | 14 | 93.33 |
| 6 | 15 | 15 | 100 | 14 | 93.33 |
| 7 | 15 | 13 | 86.67 | 13 | 86.67 |
| 8 | 15 | 14 | 93.33 | 14 | 93.33 |
| 9 | 15 | 14 | 93.33 | 14 | 93.33 |
| **Total** | **150** | **143** | **96.332** | **138** | **92** |

Table 1. Details Recognition Results

# 5. CONCLUSION

The developed system with Back-propagation Neural Network achieved the reasonable results for isolated Bangla digit speech recognition. The system can be used to recognize ten Bangla digits at a time and achieved 92% recognition accuracy for multiple speakers. With properly trained speakers and noise free environment, the developed system will produce better recognition results. The variability in various parameters, like speed, noise, and loudness will properly handle in our future research. A well-organized system should be completely speaker independent. So speakers of different ages and genders should be employed in the future researchers. In Future, speech recognition for continuous Bangla speech with more powerful recognition tools, like the Hidden Markov Model (HMM), Time Delay Neural Network (TDNN) and Gaussian Mixture Model (GMM) will be established.

**Biographies of Authors**

**Md. Ali Hossain**
Mr. Hossain was born in Manikganj, Bangladesh. He received the B.Sc. and M.Sc. degrees from the Department of Computer Science and Engineering, University of Islamic University, Kushtia, Bangladesh, in 2008 and 2009, respectively. He is serving as a Lecturer with the Department of Computer Science and Engineering (CSE), Bangladesh University, Dhaka. He has got a number of research articles published in different international journals. His current research interests include speech processing, biomedical imaging, biomedical signal, bioinformatics, neural networks and AI. Mr. Ali Hossain is an Associate Member of the Bangladesh Computer Society and Executive Member of Islamic University Computer Association (IUCA). 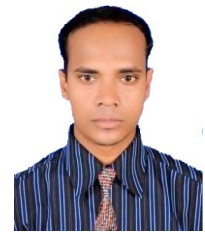

**Md. Mijanur Rahman**

Mr. Rahman is working as an assistant professor of the department of Computer Science and Engineering in Jatiya Kabi Kazi Nazrul Islam University, Trishal, Mymensingh, Bangladesh. He completed his B Sc (Hons) and M Sc in CSE degree from Islamic University, Kushtia, Bangladesh. At present he is continuing his PhD research work in the department of Computer Science and Engineering, Jahangirnagar University, Savar, Dhaka, Bangladesh. He has got a number of research articles published in different local and international journals. His current research interests include the fields of pattern recognition, image and speech processing, neural networks, fuzzy logics and AI. 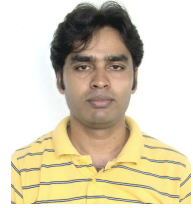

**Uzzal Kumar Prodhan**

Assistant Professor, Department of Computer Science & Engineering, Jatiya Kabi Kazi Nazrul Islam University, Trishal, Mymensingh, Bangladesh. He has completed his M.Sc. and B.Sc. from the department of Computer Science & Engineering, Islamic University, Bangladesh. He got first class in both exams. He passed S.S.C. and H.S.C with star marks. After completing M.Sc in CSE, he joined Bangladesh University as a Lecturer & joined in Jatia Kabi Kazi Nazrul Islam University as an Assistant Professor. In his long teaching life he was appointed as a head examiner in Computer Technology by Bangladesh Technical Education Board, Dhaka. Due to his teaching interest he was selected as a Book reviewer of National Curriculum of Textbook Board, Dhaka. He has successfully completed Microsoft Certified IT Professional (MCITP) on Server 2008 platform. His research interest includes Artificial Intelligence, Neural Network, Cryptography, Computer Architecture and Organization and Pattern Recognition. He has many international and national research publications. His email addresses are uzzal_bagerhat@yahoo.com, uzzal.prodhan@bu.edu.bd. 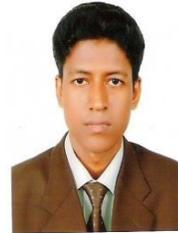

**Prof. Md. Farukuzzaman Khan**

Prof. Khan is working as professor of the department of Computer Science and Engineering in Islamic University, Kushtia, Bangladesh. He completed his B Sc (Hons), M Sc and M. Phil degree from Rajshahi University, Rajshahi, Bangladesh. He is a PhD researcher in the department of Computer Science and Engineering, Islamic University, Kushtia, Bangladesh. He has has got a number of research articles published in different local and international journals. His current research interests pattern recognition, image and speech processing, DSP and AI. 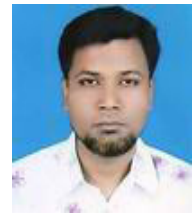